\documentclass[sigconf]{acmart}

\renewcommand\footnotetextcopyrightpermission[1]{} 

\AtBeginDocument{%
  }

\usepackage{xcolor}
\usepackage{graphicx}

\usepackage{booktabs}
\usepackage{multirow}
\usepackage{amsmath} 
\usepackage{graphicx} 

\usepackage{tabularx}

\usepackage{balance}

\begin{document}

\title{MSITrack: A Challenging Benchmark for Multispectral \\ Single Object Tracking}

\author{%
Tao Feng$^{1}$ \quad Tingfa Xu$^{1*}$ \quad Haolin Qin$^{1}$ \quad Tianhao Li$^{1}$  \quad Shuaihao Han$^{1}$ \quad 
Xuyang Zou$^{1}$ \\ \quad Zhan Lv$^{1}$ \quad Jianan Li$^{1*}$\\
$^1$Beijing Institute of Technology\\
}



\pagestyle{empty}
\fancyhead{}  
\renewcommand{\headrulewidth}{0pt} 

\begin{abstract}
  Visual object tracking in real-world scenarios presents numerous challenges including occlusion, interference from similar objects and complex backgrounds — all of which limit the effectiveness of RGB-based trackers. Multispectral imagery, which captures pixel-level spectral reflectance, enhances target discriminability. However, the availability of multispectral tracking datasets remains limited. To bridge this gap, we introduce \textbf{MSITrack}, the largest and most diverse multispectral single object tracking dataset to date. MSITrack offers the following key features: (i) \textbf{More Challenging Attributes} – including   interference from similar objects and similarity in color and texture between targets and backgrounds in natural scenarios, along with a wide range of real-world tracking challenges; (ii) \textbf{Richer and More Natural Scenes} – spanning 55 object categories and 300 distinct natural scenes, MSITrack far exceeds the scope of existing benchmarks. Many of these scenes and categories are introduced to the multispectral tracking domain for the first time; (iii) \textbf{Larger Scale} –  300 videos comprising over 129k frames of multispectral imagery. To ensure annotation precision, each frame has undergone meticulous processing, manual labeling and multi-stage verification. Extensive evaluations using representative trackers demonstrate that the multispectral data in MSITrack significantly improves performance over RGB-only baselines, highlighting its potential to drive future advancements in the field. The MSITrack dataset is publicly available at: https://github.com/Fengtao191/MSITrack.
\end{abstract}





\maketitle


\section{Introduction}

Object tracking is a fundamental problem in computer vision, with broad applications in surveillance and manufacturing. In real-world scenarios, visual trackers often encounter major challenges such as occlusion, interference from visually similar objects and low discriminability due to color or texture similarity between the target and background. These factors considerably complicate the task of visual object tracking.

\begin{figure}[h]
    \centering
    \includegraphics[width=\linewidth]{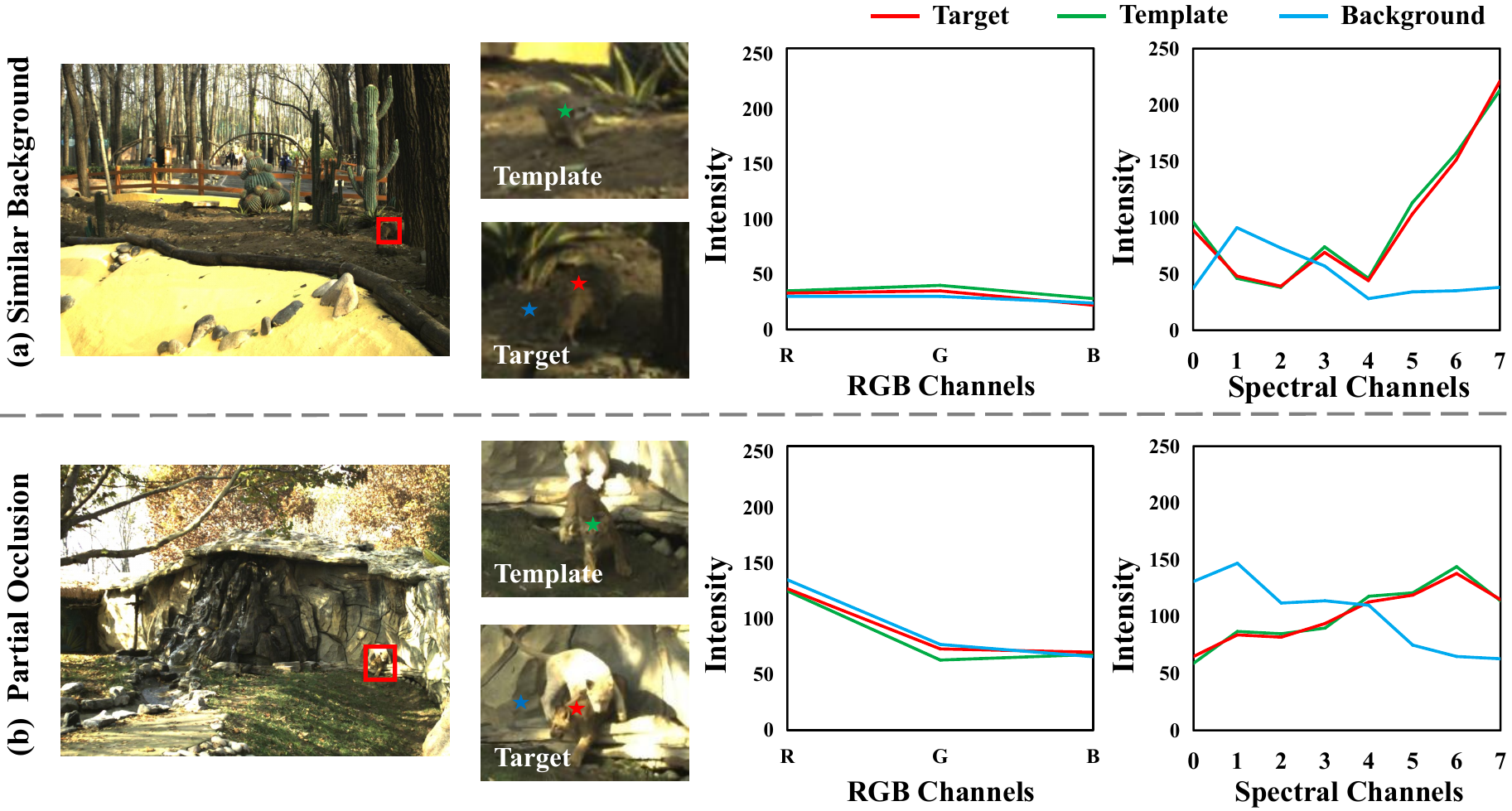}
    \caption{The target’s spectral information differs significantly from the background and aligns with the template’s spectral data, facilitating differentiation and localization.}
    \label{fig:1}
\end{figure}

\begin{table*}[h]
 \centering
  \caption{Comparison of MSITrack with other datasets. – means no relevant information was provided.}
  \label{tab:1}
  \small
  \begin{tabularx}{\textwidth}{p{2.1cm}|>{\centering\arraybackslash}p{1.2cm}cccccccc}
    \toprule
    Dataset & Modality & Channels & Task & Classes & Total Frames & Durations & Sequences & Resolution  & Wavelength Range \\
    \midrule
    OTB50\cite{wu2013online} & RGB  & 3  & General SOT& 10 & 29k & 16.4 min & 50  & - & None \\
    VOT2017\cite{kristan2016novel} & RGB & 3 & General SOT& 24 & 21k & 11.9 min & 60  & - & None \\
    TrackingNet\cite{muller2018trackingnet} & RGB & 3 & General SOT& 27  & 509k & 140 hours & >30K& - & None \\
    LaSOT\cite{fan2019lasot} & RGB & 3 & General SOT& 70 & 3.52M & 32.5 hours & 1.4K & 1280×720 & None \\
    GTOT\cite{li2016learning} & RGBT & 4 & General SOT& 9  & 7.8k & - & 50 & - & None \\
    RGBT210\cite{li2017weighted} & RGBT & 4 & General SOT& 22 & 104.7k & - & 210 & - & None \\
    MUST\cite{qin2025must} & MSI & 8 & UAV SOT& 8  & 43k & 143 min & 250 & 1200×900 & 395-950nm \\
    HOT\cite{xiong2020material} & MSI & 16 & General SOT& 20  & 21k & 14 min & 50 & 512×256 & 470-620nm \\
    \midrule
    \noindent\textbf{MSITrack} (Ours) & MSI & 8 & General SOT& 55 & 129k & 430 min & 300 & 1200×900 & 395-950nm \\
  \bottomrule
\end{tabularx}
\end{table*}

Existing RGB-based tracking algorithms\cite{ye2022joint,wu2023dropmae,kou2023zoomtrack} primarily rely on spatial appearance features—such as color and texture—to detect and track objects. However, in complex scenarios involving occlusion or background similarity, these appearance cues can degrade severely, compromising the algorithm’s ability to reliably distinguish and track the target. As illustrated in Fig.~\ref{fig:1}(a), a meerkat is nearly indistinguishable from the background, demonstrating the limitations of appearance-based tracking methods. This highlights the need to explore feature dimensions beyond visual appearance in order to enhance object distinguishability and improve the accuracy and robustness of general-purpose tracking algorithms under real-world conditions.

Multispectral imaging (MSI) captures the spectral reflectance of objects at the pixel level across a wide range of wavelengths, thereby providing richer information on material composition and surface properties information\cite{qin2024dmssn,islam2024hy,gao2023cbff} than RGB imagery. In challenging scenarios involving background interference or occlusion, MSI provides complementary spectral cues that preserve target identity even when spatial features are compromised. As shown in Fig.~\ref{fig:1}(b), a distinct spectral curve allows for effective target-background separation, even under severe occlusion. Compared to RGB data, multispectral imagery offers more discriminative spectral signatures\cite{liang2018material}, presenting a more robust solution for object tracking. However, progress in this area is hindered by the lack of large-scale, high-quality multispectral tracking datasets.

To address this limitation, we present MSITrack—the largest and most diverse multispectral single object tracking dataset. Collected using a snapshot multispectral camera, MSITrack captures real-world scenes under a wide range of temporal and environmental conditions. The dataset features three major characteristics:




\noindent\textbf{(i) More Challenging Attributes.} The collected sequences are intentionally designed to be highly challenging, capturing scenarios such as interference from similar objects and visual similarity between the target and background in terms of color and texture. These challenges closely mirror real-world conditions.
    
\noindent\textbf{(ii)Richer and More Natural Scenes.} To enhance the diversity and generalization capabilities of multispectral tracking, MSITrack includes 55 object  categories  and 300 unique natural scenes — substantially more than existing benchmarks such as HOT (20 categories and 21 scenes). Many of the newly introduced categories and scenes are appearing in the context of multispectral tracking for the first time, thereby expanding the coverage and applicability.
    
\noindent\textbf{(iii) Larger Scale.} The dataset consists of 300 videos comprising a total of 129k multispectral frames, each with a spatial resolution of 1200 × 900 pixels. It spans eight spectral bands ranging from 395 nm to 950 nm, covering the visible to near-infrared spectrum. These sequences represent 300 distinct natural scenes. All frames have been manually annotated through a meticulous process requiring over 1,300 labor hours, ensuring high-quality bounding boxes and providing a robust foundation for reliable evaluation.

The proposed dataset provides a solid foundation for advancing the field of multispectral single object tracking. To benchmark existing methods, we conducted a comprehensive evaluation of a wide range of recent state-of-the-art trackers. Experimental results consistently show that trackers leveraging multispectral input significantly outperform their RGB-based counterparts, particularly in complex scenarios where spatial appearance features are limited. These findings underscore the potential of MSITrack to support and inspire future advancements in this domain.

Our main contributions are as follows: (i) We present MSITrack, the largest and most diverse dataset for multispectral single object tracking to date; (ii) The dataset incorporates more challenging attributes and support the development of more robust tracking algorithms; (iii)We perform extensive experimental evaluations, establishing strong baselines to guide future research and comparative studies. All dataset resources and source code are publicly available to encourage further development and promote reproducibility.

\begin{figure}[t]
    \centering
    \includegraphics[width=\linewidth]{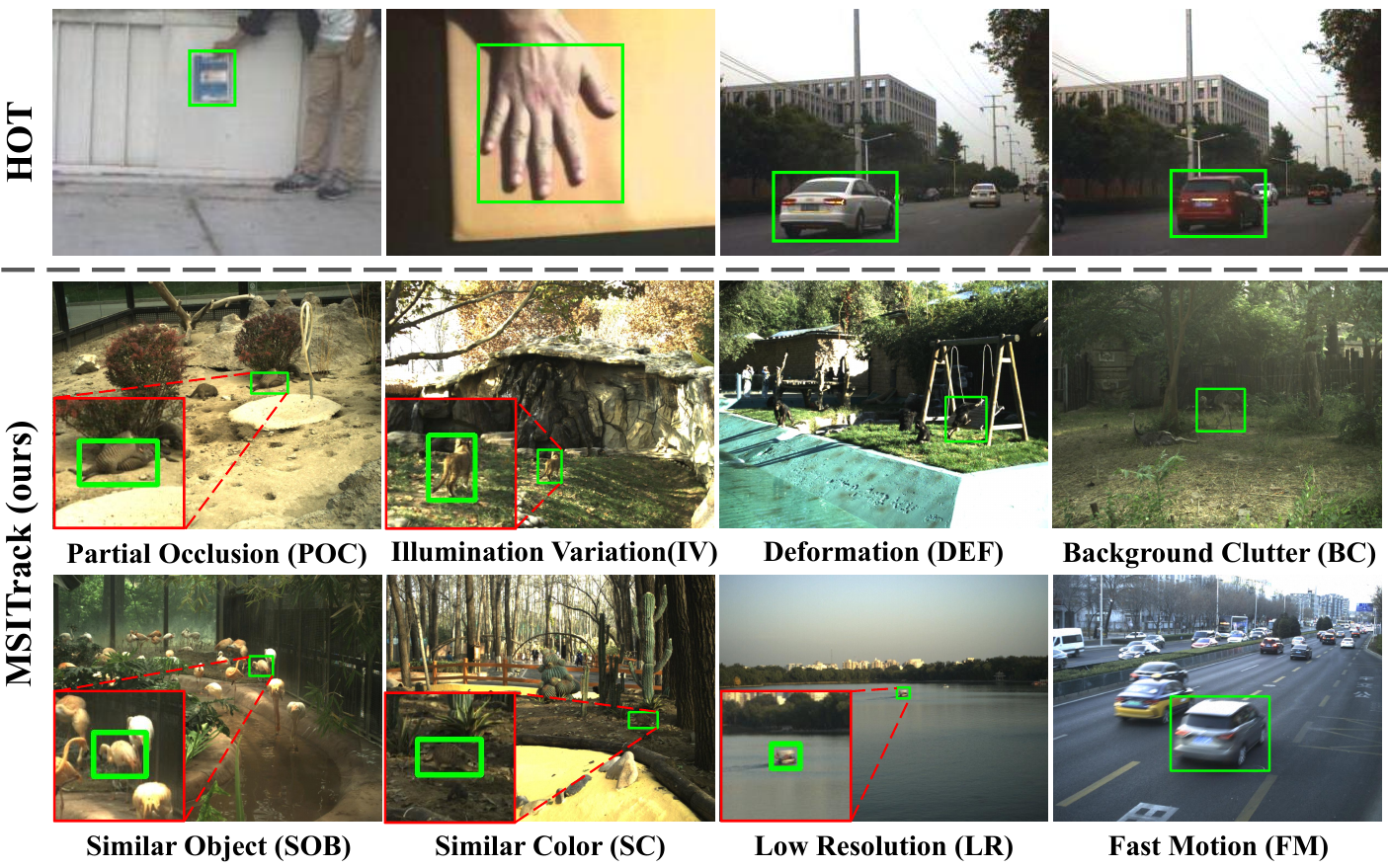}
    \caption{Comparison of scenes between MSITrack and HOT.}
    \label{fig:2}
\end{figure}

\begin{figure*}[t]
    \centering
    \includegraphics[width=\linewidth]{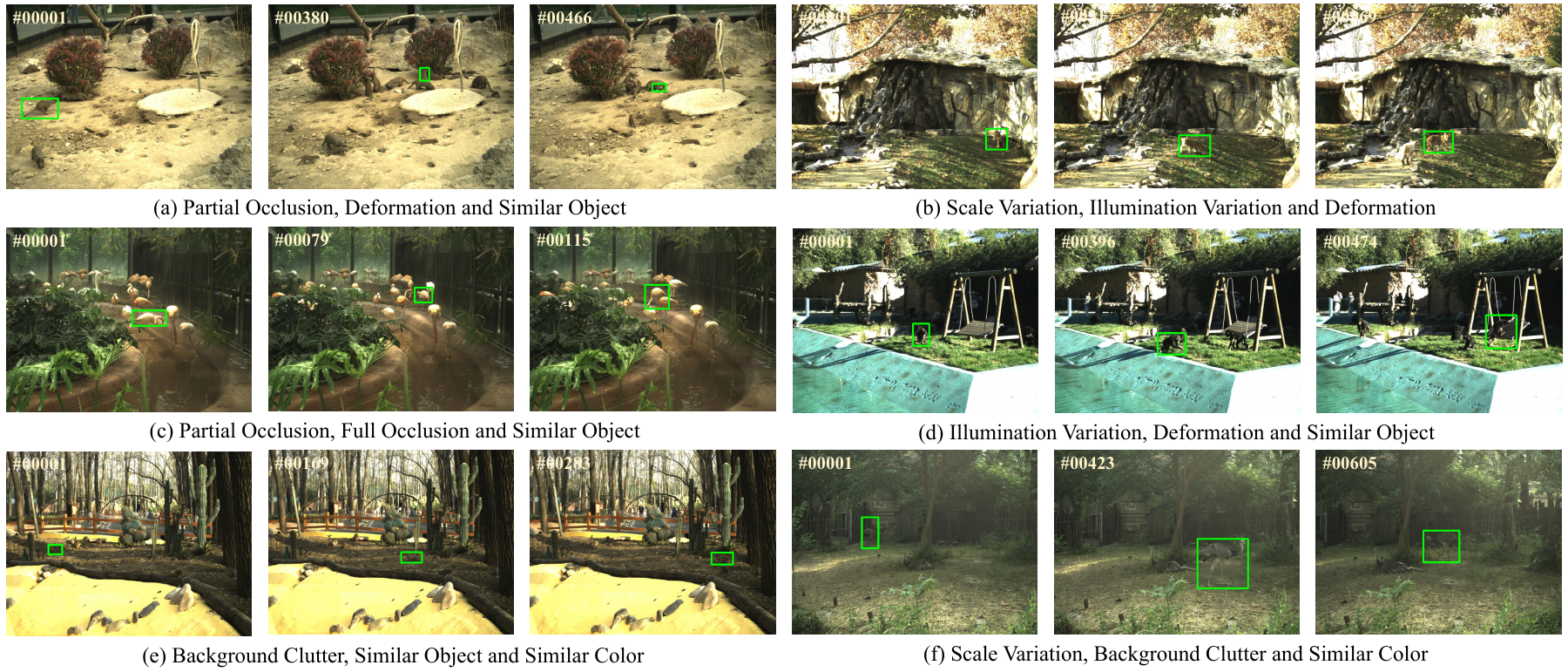}
    \caption{Visualization of several annotation examples in the proposed MSITrack.}
    \label{fig:3}
\end{figure*}

\section{Related Work}

\noindent\textbf{MSI-based Tracking Datasets.}
Several MSI datasets have recently been introduced for visual object tracking. The HOT\cite{xiong2020material} dataset is a ground-based collection acquired using a snapshot mosaic hyperspectral camera, comprising 50 videos across 20 object categories. Due to hardware limitations, its spectral coverage is confined to the 470–620 nm range, with a spatial resolution of 512 × 256 pixels. The MUST\cite{qin2025must} dataset, by contrast, is designed specifically for aerial-view multispectral tracking. It includes 250 videos spanning 8 object categories and demonstrates the benefits of spectral data in airborne environments. All scenes in MUST are captured from a high-altitude top-down perspective over urban streets, resulting in small target sizes due to the elevated viewpoint. While the aforementioned datasets have significantly contributed to the progress of multispectral tracking, their overall scale remains relatively limited. To overcome this limitation, we propose MSITrack, a challenging dataset encompassing a broader and more diverse set of scenes.

\noindent\textbf{RGB-based Tracking Datasets.}
Early tracking datasets were limited both in variety and scale. OTB-2013\cite{wu2013online}, the first single object tracking dataset, contained only 50 videos and 10 categories. TC-128\cite{liang2015encoding} consists of 128 RGB videos, while VOT2017\cite{kristan2016novel} introduced a series of challenges to better compare the performance of various trackers. TrackingNet\cite{muller2018trackingnet} proposed a large-scale tracking dataset, including over 30k tracking sequences. LaSOT\cite{fan2019lasot} contains 1,400 long sequences, and GOT-10k\cite{huang2019got} expanded the tracking categories to 563 categories and 10k videos. In contrast, MSITrack introduces a complementary and discriminative spectral dimension that offers a more effective solution for object tracking.

\noindent\textbf{RGBT-based Tracking Datasets. }
The GTOT\cite{li2016learning} dataset comprises 50 RGBT videos, using visible light and thermal infrared images for tracking. Li et al.\cite{li2017weighted} introduced the RGBT210 dataset, which includes 210 RGBT videos featuring annotations for 12 challenge attributes. This dataset was subsequently expanded into RGBT234\cite{li2019rgb}, which contains 234 RGBT videos, offering improved annotation quality and broader challenge coverage. However, thermal infrared imaging is susceptible to interference from external heat sources\cite{li2021lasher,zhang2019multi,lan2020modality}. In contrast, multispectral imaging demonstrates stable performance under natural daylight conditions and is less prone to errors caused by thermal variations\cite{hwang2015multispectral,yu2021feedback,uzair2015hyperspectral}.


\section{MSITrack Dataset}

\begin{figure*}[h]
    \centering
    \includegraphics[width=\linewidth]{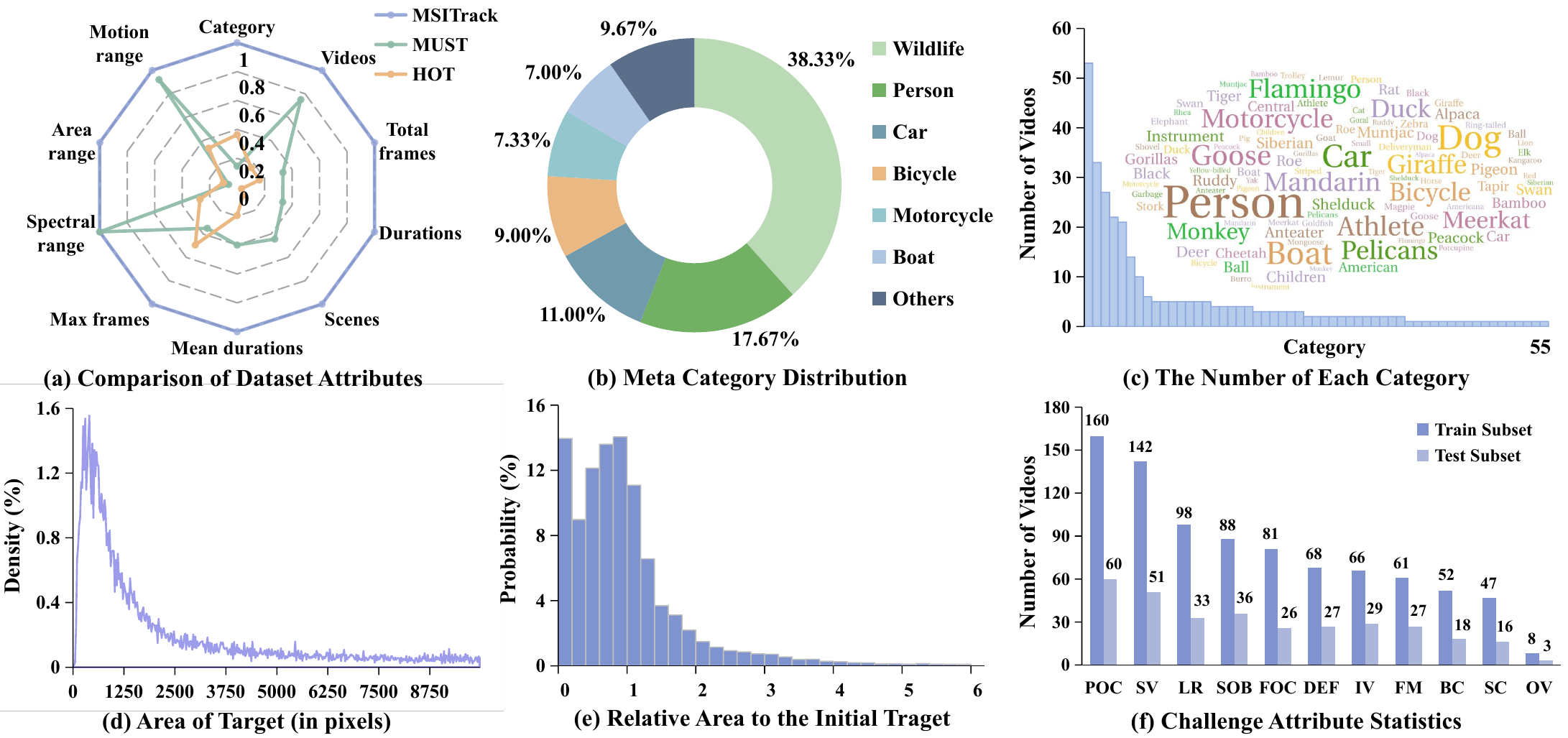}
    \caption{Comparison of datasets, category distribution, object size, relative area and dataset splitting in MSITrack.}
    \label{fig:4}
\end{figure*}

\subsection{Construction Principle}

MSITrack is designed to serve as a comprehensive, large-scale and highly challenging benchmark tailored for general object tracking in real-world scenarios. To this end, the construction of MSITrack follows several key principles:

\noindent\textbf{• More Complex Real-World Challenges.} The dataset should capture extensive variations in  background complexity, occlusion levels, and include more challenging attributes such as similar objects interference, color similarity between targets and backgrounds, offering comprehensive coverage of the real-world challenges.

\noindent\textbf{• More Diverse and Real-World Scenes.} A rich variety of tracking categories and natural scenes is critical for advancing generalizable multispectral tracking. MSITrack should encompass a variety of dynamic environments, including urban areas, forests, wetlands and lakes, and include more than 50 object categories, further broadening its applicability and relevance.

\noindent\textbf{• Larger Scale. } A primary goal of MSITrack is to deliver a large-scale multispectral dataset that supports both the training of deep learning models and robust performance evaluation. To this end, we target a large-scale dataset comprising 300 videos and over 100k multispectral images, enabling general-purpose object tracking across a broad range of conditions.

\subsection{Dataset Overview}
As shown in Fig.~\ref{fig:4}(a), MSITrack is currently the largest and most diverse multispectral single object tracking dataset, comprising 300 videos and over 129k frames of multispectral imagery. As illustrated in Fig.~\ref{fig:4}(c), MSITrack includes a wide range of tracking categories. In addition to commonly seen classes such as pedestrian, car and bicycle, the dataset features less frequently encountered objects such as meerkat, lion, gorilla, flamingo, elk, monkey and giraffe, totaling 55 distinct categories. Many of these classes are introduced to the multispectral tracking domain for the first time.

Tab.~\ref{tab:1} presents a comparative overview of MSITrack and several representative general-purpose and multispectral single object tracking datasets. The MUST dataset targets traffic-related objects from an aerial, top-down viewpoint. However, its limited range of object categories constrains its utility in broader visual tracking applications. HOT, one of the earliest general-purpose multispectral tracking datasets, has significantly contributed to the development of the field. Nevertheless, as shown in Fig.~\ref{fig:2}, HOT primarily includes relatively simple targets such as books, hands and cars, some of which are manually manipulated, and its scenes often overlap. 

In contrast, MSITrack spans a significantly broader spectral range—from 395 nm to 950 nm—covering wavelengths from the visible to the near-infrared spectrum. It also marks a substantial increase in scale, with a greater variety of object categories (55 vs. 20) and more diverse, natural scenes (300 vs. 21), each of which is unique. Importantly, MSITrack emphasizes real-world complexity by including a substantial number of videos featuring similar-object interference (SOB) and low target-to-background separability in terms of color and texture (SC). These challenging attributes more precisely capture the complexities of real-world tracking scenarios involving heterogeneous object types and diverse environments, thereby facilitating further research in this domain.

\begin{table}[t]
  \caption{Descriptions of 11 different attributes in MSITrack}
  \label{tab:attribute}
  \normalsize
  \begin{tabularx}{\linewidth}{@{}p{1.0cm}ll@{}}
    \toprule
    Attribute&Definition\\
    \midrule
    POC & the target object is partially occluded in the sequence\\
    SV & the ratio of bounding box is outside [0.5, 2]\\
    LR & the area of target box is smaller than 1000 pixels\\
    SOB & the target is surrounded by visually similar distractors\\
    FOC & the target object is fully occluded in the sequence \\
    DEF & the target object is deformable during tracking\\
    IV & the illumination in the target region changes\\
    FM & the motion of target object is larger than 50 pixels\\
    BC & the background of the target has some clutter\\
    SC & the background has a color similar to that of the target\\
    OV & the target object completely leaves the video frame\\
  \bottomrule
\end{tabularx}
\end{table}

\subsection{Dataset Construction}

\noindent\textbf{Data Acquisition. } MSITrack was developed using a snapshot-based multispectral camera capable of capturing images across eight spectral bands, covering both the visible and near-infrared ranges, with a spectral sensitivity from 395 nm to 950 nm. To ensure the dataset's relevance to general-purpose object tracking tasks, data collection was conducted under diverse weather conditions, including sunny, overcast and foggy days.

The dataset spans a broad range of environments, such as urban streets, lakes and wetlands, forest parks, sports fields, playgrounds and wildlife reserves. Initially, over 290k raw frames were captured. This careful curation process resulted in a final dataset of 129k frames, forming 300 tracking sequences with an average length of 431 frames per sequence. All frames underwent precise alignment procedures, including both geometric and radiometric corrections, to ensure the generation of high-quality multispectral sequences optimized for general-purpose object tracking.

\noindent\textbf{Annotation. } MSITrack is a carefully curated dataset containing over 1300 human-hours of manual annotations. Bounding box annotations follow a rigorous set of guidelines\cite{peng2024vasttrack}. If the target is fully occluded or out of the frame, no bounding box is provided. We established an annotation team, referred to as LYN, comprising a small group of experts and ten qualified annotators with experience in object tracking. The annotation process was organized into two stages—annotation and verification—implemented through a multi-step workflow that included manual labeling, visual inspection and cross-validation.

In the first stage, after an expert annotated the initial frame of each sequence, the remaining frames were manually annotated by the trained annotators using a standardized annotation tool. Upon completion, the process advanced to the verification stage, where three experienced annotators from the original team conducted cross-validation and refined the annotations to ensure accurate bounding box placement and sizing. Each bounding box was meticulously adjusted to tightly fit the target with minimal margin. Fig.~\ref{fig:3} shows several annotation examples.




\begin{table}[t]
  \caption{Detailed statistics across datasets. Scenes: Non-repeating scenes; Time: Average duration; Frames: Maximum number of frames; Motion: Maximum moving pixels.}
  \label{tab:2}
  \begin{tabularx}{\linewidth}{l|>{\centering\arraybackslash}X>{\centering\arraybackslash}X>{\centering\arraybackslash}X>{\centering\arraybackslash}X}
    \toprule
    Dataset& Scences  & Time  & Frames & Motion\\
    \midrule
    MUST & 133  & 34s  & 790 & 453\\
    HOT & 21   & 17s & 1111  & 16\\
    \midrule
    \textbf{MSITrack} & \textbf{300}   & \textbf{86s}  & \textbf{2251} & \textbf{492}\\
  \bottomrule
\end{tabularx}
\end{table}

\noindent\textbf{Dataset Splitting. }MSITrack is divided into a training set and a test set. To ensure fairness and promote generalization, environmental factors—such as weather conditions—are evenly distributed across both subsets. The final split comprises 220 training sequences and 80 testing sequences. The training set contains 94k frames, with an average sequence length of 431 frames, while the test set includes 35k frames, with an average sequence length of 429 frames. As illustrated in Fig.~\ref{fig:4}(f), we carefully maintain an even distribution of 11 key challenge attributes across both the training and testing sets, thereby ensuring comparable difficulty and diversity between the subsets. 







\subsection{Statistical Analysis}

\noindent\textbf{Challenging Attributes. } As
shown in Tab.~\ref{tab:attribute}, MSITrack incorporates annotations for 11 challenging tracking attributes across all sequences. As illustrated in Fig.\ref{fig:4}(f), partial occlusion is the most frequently encountered challenge, present in 220 videos. More complex conditions—such as similar-object interference and target-background color similarity—occur in 124 sequences (41.3\%) and 63 sequences (21.0\%), respectively. The former evaluates a tracker's ability to preserve target identity under object ambiguity, while the latter assesses sensitivity to subtle appearance differences and fine-grained target discrimination. As shown in Tab.~\ref{tab:2}, MSITrack offers a substantially greater diversity of unique scenes compared to existing datasets. These attribute-driven analyses highlight the complexity of the MSITrack, limiting the effectiveness of purely spatial feature approaches and motivating the development of more robust and adaptive tracking algorithms.

\noindent\textbf{Category and Target Size Distribution. } As shown in Fig.~\ref{fig:4}(b) and Fig.~\ref{fig:4}(c), MSITrack includes a diverse array of target categories, with the Wildlife meta-category comprising the largest proportion. Wildlife targets typically exhibit a wide range of fur textures, colors and body structures, thereby enriching the dataset’s spectral feature space and enhancing its relevance to real-world scenarios. In contrast to rigid objects such as vehicles, animal targets present substantial non-rigid deformations and irregular motion patterns, and are often situated in visually complex and dynamically changing environments. These attributes collectively support research in fine-grained target discrimination and robustness to deformation under practical conditions. The target area distribution depicted in Fig.~\ref{fig:4}(d) further substantiates this observation, with the primary peak occurring around medium-sized targets (about 1000 pixels), and a long right tail indicating the inclusion of large-scale targets.

\noindent\textbf{Trajectory Duration and Inter-Frame Motion Analysis. } As summarized in Tab.~\ref{tab:2}, MSITrack features significantly longer average sequence durations—2.5 times those of MUST and 5.1 times those of HOT. Additionally, MSITrack surpasses existing multispectral datasets in both the maximum frame count per sequence and the maximum displacement of targets. The distribution
of relative area to the initial traget, shown in Fig.~\ref{fig:4}(e), reflects realistic and diverse scale variations, which are commonly encountered in practical tracking scenarios. These attributes introduce greater challenges for tracking algorithms, particularly in terms of responsiveness and robustness, thereby enabling a more comprehensive evaluation of their long-term stability and temporal consistency.



\section{Experiments}

\subsection{Experimental Settings}

We conducted algorithm evaluations on the MSITrack dataset using two types of input: RGB and MSI. For the RGB-based evaluation, we utilized images captured by the RGB camera of the multispectral system, synchronized and aligned to the same viewpoint. For the MSI-based evaluation, data from all eight spectral channels were employed, only changing the number of input channels to the tracker from 3 to 8 during training and testing. Due to the lack of available pre-trained parameters for MSI, we instead initialize the network with ImageNet-trained weights\cite{deng2009imagenet,he2022masked} and employ a simple yet effective parameter reconstruction strategy. The strategy aims to extend RGB-based pre-trained parameters to MSI-based vision tasks through replication.

Following\cite{wu2013online,fan2019lasot,muller2018trackingnet}, we use one-pass evaluation (OPE) and compare different trackers on MSITrack using five metrics: Area Under the Curve (AUC) of the success plot, SR\textsubscript{0.5}, SR\textsubscript{0.75}, precision (PRE) and normalized precision (NPRE).

\begin{table}[t]
  \caption{Comparison with state-of-the-art trackers on MSITrack. The top two results are highlighted in \textcolor{red}{red} and \textcolor{blue}{blue}.}
  \label{tab:3}
  \begin{tabularx}{\linewidth}{l|XXXXX}
    \toprule
    \textbf{Method} & \textbf{AUC} & \textbf{SR$_{0.5}$} & \textbf{SR$_{0.75}$} & \textbf{Pre} & \textbf{Pre$_N$}\\
    \midrule
    SGLATrack\cite{xue2025similarity}  & 35.9 & 43.7 & 27.6 & 48.5 & 43.9\\
    HIPTrack\cite{cai2024hiptrack}  & 36.2 & 43.1 & 29.0 & 47.8 & 43.0\\
    OSTrack$_{256}$\cite{ye2022joint}  & 42.0 & 50.9 & 34.2 & 55.7 & 50.5\\
    OSTrack$_{384}$\cite{ye2022joint}  & 43.1 & 52.5 & 36.1 & 56.2 & 52.4\\
    LMTrack\cite{xu2025less}  & 43.2 & 52.4 & 39.7 & 55.2 & 52.5\\
    ZoomTrack\cite{kou2023zoomtrack}  & 44.3 & 54.7 & 35.1 & 59.0 & 54.4\\
    NeighborTrack\cite{chen2023neighbortrack}  & 44.8 & 55.2 & 35.7 & 59.7 & 54.6\\
    EVPTrack\cite{shi2024explicit}  & 46.2 & 55.8 & \textbf{\textcolor{blue}{43.9}} & 58.3 & 56.0\\
    AQATrack\cite{xie2024autoregressive}  & 47.1 & 56.7 & \textbf{\textcolor{red}{44.7}} & 58.5 & 56.9\\
    UNTrack\cite{qin2025must}  & \textbf{\textcolor{blue}{47.3}} & \textbf{\textcolor{red}{58.4}} & 40.5 & \textbf{\textcolor{blue}{61.8}} & \textbf{\textcolor{blue}{57.8}}\\
    GRM\cite{gao2023generalized}  & \textbf{\textcolor{red}{47.8}} & \textbf{\textcolor{blue}{58.1}} & 40.8 & \textbf{\textcolor{red}{63.0}} & \textbf{\textcolor{red}{57.8}}\\
    
  \bottomrule
  
\end{tabularx}
\end{table}

\begin{figure}[t]
    \centering
    \includegraphics[width=\linewidth]{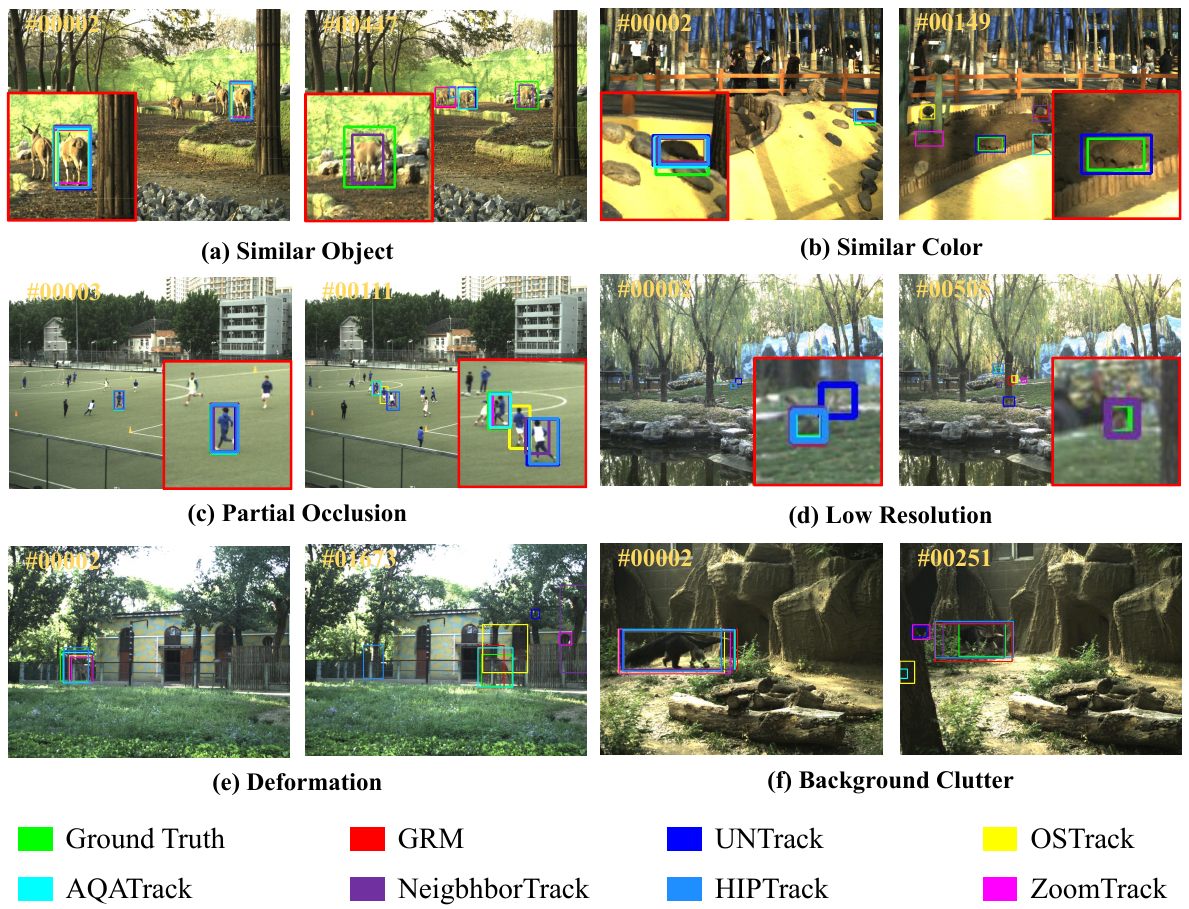}
    \caption{Comparison of some state-of-the-art trackers when dealing with various challenge attributes.}
    \label{fig:5}
\end{figure}

\subsection{Experimental Results and Analysis}

\noindent\textbf{Overall Performance. } All models were evaluated under comprehensive and fair conditions, with detailed results presented in Tab.~\ref{tab:3}. Among the state-of-the-art trackers assessed, GRM achieved the best performance across all metrics, with scores of 0.478 AUC, 0.630 PRE and 0.578 NPRE. UNTrack ranked second overall, attaining scores of 0.473 AUC, 0.618 PRE and 0.578 NPRE. The third-ranked tracker, AQATrack, achieved scores of 0.471 AUC, 0.585 PRE and 0.569 NPRE. Notably, UNTrack is specifically designed for multispectral object tracking. It employs a spectral background removal module to aggregate useful spectral information while narrowing the search area, underscoring the significance of spectral information in deep tracking. The tracking results of each algorithm are visualized in Fig.~\ref{fig:5}.


\noindent\textbf{Benefits of Multispectral Cues. }
To further quantify the advantages of spectral information in object tracking, we compared the performance of the same tracking algorithms using RGB input. As shown in Tab.~\ref{tab:4}, algorithms utilizing MSI input demonstrated significant improvements across all five performance metrics. Specifically, GRM’s AUC score increased by 5.3\%, AQATrack’s AUC improved by 3.7\%, and OSTrack256’s AUC rose by 3.3\%. These results underscore the effectiveness of spectral information in enhancing target representation under more challenging scenarios.

\begin{table}[t]
\caption{Comparison between RGB and MSI Input.}
\centering
\begin{tabularx}{\linewidth}{l|c|>{\centering\arraybackslash}p{1.8cm}XX}
\toprule
\textbf{Method} & \textbf{Input} & \textbf{AUC}  & \textbf{Pre} & \textbf{Pre$_N$} \\
\midrule
\multirow{2}{*}{GRM\cite{gao2023generalized}} 
    & RGB & 42.5  & 55.2 & 51.8 \\
    & \textbf{MSI} & \textbf{47.8~($\uparrow$5.3)}  & \textbf{63.0} & \textbf{57.8} \\
    \midrule
\multirow{2}{*}{AQATrack\cite{xie2024autoregressive}} 
    & RGB & 43.4  & 56.2 & 52.5 \\
    & \textbf{MSI} & \textbf{47.1~($\uparrow$3.7)}  & \textbf{58.5} & \textbf{56.9} \\
\midrule
\multirow{2}{*}{OSTrack$_{256}$\cite{ye2022joint}} 
    & RGB & 38.7  & 51.0 & 46.8 \\
    & \textbf{MSI}  & \textbf{42.0~($\uparrow$3.3)}  & \textbf{55.7} & \textbf{50.5} \\
\bottomrule

\end{tabularx}
\label{tab:4}
\end{table}



\begin{table}[t]
\caption{A comparison of three state-of-the-art trackers in AUC across each challenge attribute.}
\vspace{-2mm}
\centering
\begin{tabularx}{\linewidth}{l|c|XXXXXXX}
\toprule
\textbf{Method} & \textbf{Input}  & \textbf{LR} & \textbf{FOC}& \textbf{IV} & \textbf{SC}& \textbf{SV}& \textbf{POC} & \textbf{SOB} \\
\midrule
\multirow{2}{*}{GRM} 
    & RGB   & 40.4 & 35.8 & 42.1  & 40.7 & 44.7 & 39.9 & 34.8 \\
    & \textbf{MSI}   & \textbf{48.5} & \textbf{43.5} & \textbf{50.3}  & \textbf{46.9}  & \textbf{50.3} & \textbf{43.4}& \textbf{39.0} \\
    \midrule
\multirow{2}{*}{AQATrack} 
    & RGB   & 40.3 & 36.7 & 43.4  & 42.5 & 43.9 & 40.8 & 33.8\\
    & \textbf{MSI}   & \textbf{45.6} & \textbf{40.0} & \textbf{48.0}  & \textbf{45.1} & \textbf{48.5} & \textbf{43.2}& \textbf{36.7}\\
\midrule
\multirow{2}{*}{OSTrack$_{256}$} 
    & RGB   & 35.3 & 28.1 & 40.5  & 35.8  & 40.0 & 33.4 & 31.3\\
    & \textbf{MSI}    & \textbf{42.9} & \textbf{36.1} & \textbf{44.9}  & \textbf{42.9} & \textbf{44.1} & \textbf{37.9} & \textbf{33.1} \\
\bottomrule

\end{tabularx}
\label{tab:attribute AUC}
\vspace{-3mm}
\end{table}

In addition, we compared the performance of trackers using RGB and MSI inputs across various challenging attributes, as shown in Tab.~\ref{tab:attribute AUC}. Trackers with MSI input consistently outperformed their RGB counterparts in nearly all challenging scenarios. Spectral information demonstrated remarkable effectiveness in handling cases where visual appearance is severely compromised—such as low resolution (LR) and full occlusion (FOC)—achieving average AUC improvements of 7.0\% and 6.3\%, respectively, across the three trackers. Under illumination variation (IV), the average AUC gain reached 5.7\%. In such conditions, the spatial appearance features relied upon by RGB inputs tend to degrade or become ambiguous, thereby reducing the distinguishability of the target. In contrast, the spectral features provided by MSI offer complementary cues that significantly enhance target separability. Notably, in more complex scenarios such as similar colors (SC) and similar object (SOB), the unique spectral signatures contributed to average AUC improvements of 5.3\% and 2.9\%, respectively. These findings highlight the critical role of spectral information in supporting robust object tracking under diverse and challenging conditions.

\begin{figure}[t]
    \centering
    \includegraphics[width=\linewidth]{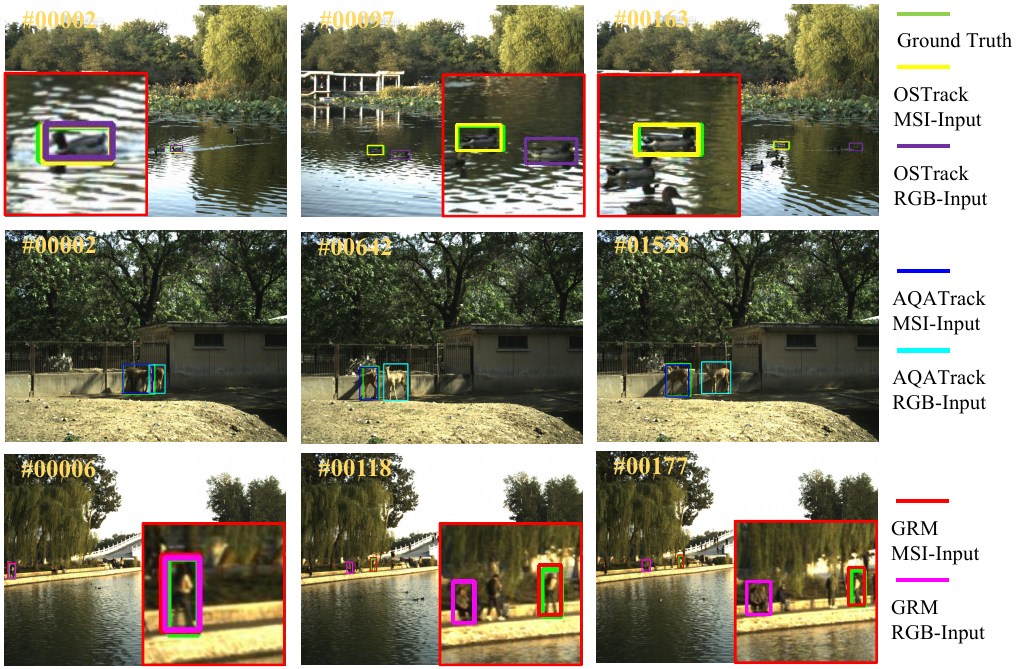}
    \vspace{-6mm}
    \caption{Comparison of three representative trackers (OSTrack, AQATrack and GRM ) using RGB and MSI inputs.}
    \label{fig:7}
    \vspace{-5mm}
\end{figure}

\noindent\textbf{Qualitative Comparison of RGB and MSI Inputs. }
As shown in Fig.~\ref{fig:7}, multispectral input significantly enhances tracking reliability under visually challenging conditions. In the top row, the RGB-based OSTrack (purple box) fails to correctly follow the target mandarin duck due to interference from visually similar objects. In the second row, RGB-based AQATrack (cyan box) misidentifies the target elk in a scene affected by both illumination changes and similar-object interference. In the third row, RGB-based GRM (magenta box) struggles to effectively separate the pedestrian from the background in low-resolution with varying lighting conditions. In contrast, the MSI-based trackers demonstrate more accurate target identification and more stable tracking performance. These qualitative results are not isolated incidents but rather represent consistent patterns observed across the dataset. They highlight how multispectral input improves target discriminability, reduces misidentification and enhances overall tracking robustness in complex scenarios.

\section{Conclusion}

We present MSITrack, the largest and most comprehensive multispectral single-object tracking dataset to date, consisting of 300 videos and 129k multispectral images, spanning 55 object categories. MSITrack introduces greater tracking challenges by incorporating numerous scenes with visually similar distractors and low-saliency targets, thereby facilitating the development of models with enhanced robustness and background suppression capabilities in real-world environments. Extensive experiments on 11 state-of-the-art trackers consistently demonstrate that incorporating spectral inputs yields significant performance improvements, especially for small or densely clustered objects where spatial cues are limited. All data and code are publicly available to support further research in this domain.

\noindent\textbf{Limitation. }
High-quality annotated tracking bounding boxes require substantial manual effort. Future work will explore scalable, unsupervised annotation methods, such as language supervision and unsupervised learning techniques.

\bibliographystyle{ACM-Reference-Format}
\balance
\bibliography{main}










\end{document}